\documentclass[10pt,sigconf,letterpaper]{acmart}
\usepackage[utf8]{inputenc}
\usepackage{amsmath,amssymb}
\usepackage{graphicx}
\usepackage{enumitem}
\usepackage{xcolor}
\usepackage{hyperref}
\usepackage{booktabs}
\usepackage{tabularx}
\usepackage{tikz,graphics,color,float,epsf,caption,subcaption}
\usepackage{balance}

\setcopyright{none}
\renewcommand\footnotetextcopyrightpermission[1]{} % removes footnote with conference information in first column
\begin{document}
\acmYear{}
\copyrightyear{}
\setcopyright{acmcopyright}
\acmConference{Unsupervised XAI}{Clustering}{Network Measurements}
\acmPrice{}
\acmDOI{}
\acmISBN{}

\title[EXPLAIN-IT: XAI for Unsupervised Network Traffic Analysis]{EXPLAIN-IT: Towards Explainable AI\\for Unsupervised Network Traffic Analysis}

\author{Andrea Morichetta}
\affiliation{
  \institution{Politecnico di Torino\\AIT Austrian Institute of Technology}
}
\email{andrea.morichetta@polito.it}

\author{Pedro Casas}
\affiliation{
  \institution{AIT Austrian Institute of Technology}
}
\email{pedro.casas@ait.ac.at}

\author{Marco Mellia}
\affiliation{
  \institution{Politecnico di Torino}
}
\email{marco.mellia@polito.it}

% \author{Andrea Morichetta$^{*,\dagger}$, Pedro Casas$^{\dagger}$, Marco Mellia$^*$}
% \affiliation{$*$ Politecnico di Torino, $\dagger$ AIT Austrian Institute of Technology}
% \email{andrea.morichetta@polito.it -- pedro.casas@ait.ac.at -- marco.mellia@polito.it}

\renewcommand{\shortauthors}{A. Morichetta, P. Casas, M. Mellia}

\begin{abstract}
The application of unsupervised learning approaches, and in particular of clustering techniques, represents a powerful exploration means for the analysis of network measurements. Discovering underlying data characteristics, grouping similar measurements together, and identifying eventual patterns of interest are some of the applications which can be tackled through clustering. Being unsupervised, clustering does not always provide precise and clear insight into the produced output, especially when the input data structure and distribution are complex and difficult to grasp. In this paper we introduce EXPLAIN-IT, a methodology which deals with unlabeled data, creates meaningful clusters, and suggests an explanation to the clustering results for the end-user. EXPLAIN-IT relies on a novel explainable Artificial Intelligence (AI) approach, which allows to understand the reasons leading to a particular decision of a supervised learning-based model, additionally extending its application to the unsupervised learning domain. We apply EXPLAIN-IT to the problem of YouTube video quality classification under encrypted traffic scenarios, showing promising results.
\end{abstract}
\maketitle

\section{Introduction}\label{sec:intro}

The undeniable popularity of Artificial Intelligence, and in particular of Machine Learning (ML) techniques, have also concerned the network community in the last decade \cite{nguyen2008surveyML, boutaba_comprehensive_2018}. The capability of addressing big data problems and automating processing measurements is appealing for multiple problems, from network security to Quality of Experience (QoE) monitoring and analysis \cite{boutaba_comprehensive_2018}.

When it comes to ML techniques and methodologies, we often and more extensively refer to supervised approaches. Supervised learning builds a model starting from the data, requiring these to be apriori categorized, i.e., labeled according to the ground truth. Ground truth is generally missing due to the structural complexity of the data, limits of human knowledge, and significant volumes that complicate the categorization process. This scenario is especially critical when it comes to network traffic, where researchers and practitioners have indeed to deal with small and outdated datasets.

Unsupervised techniques offer a solution to this lack of ground-truth since their goal is to analyze the structural properties of the data, based on some form of similarity among data instances. Different approaches are possible, depending on the overall goal (e.g., outlier detection, categorization, and others), and the different levels of complexity of the analysis. In any case, there is no need for ground truth. However, analyzing and interpreting the results obtained through clustering is a cumbersome and challenging task, often requiring time and sophisticated, expert-based manual inspection. In most cases, the complexity of the data - volume and dimensionality, and the size of the obtained results - number of clusters and outliers, is such that manual inspection analysis becomes forbidding. Unsupervised quality metrics, such as the Silhouette or Rank Index \cite{iglesias2019}, provide only structural insights on the obtained results, but they do not explain why the clustering methodology grouped points in the same cluster. Supervised quality metrics, such as cohesiveness and purity, require ground truth and, even when it is available, this approach suffers if ground truth is partial, wrong, or biased. 

When it comes to the interpretability of results, traditional approaches are based on \emph{white-box} supervised techniques to explain the decisions provided by a particular ML model. White-box techniques use simple and easy-to-interpret models such as linear discriminant functions or decision trees, which generally offers a straightforward interpretation of the models' decisions. However, white box explainability limits the set of applicable algorithms to those who are natively interpretable, with the additional drawback of potentially limiting performance.

Recent work and efforts in the field of Explainable Artificial Intelligence (XAI) \cite{gunning2017explainable, miller2019explanation, hazard2019} provide both useful terminology and discussion on where and how such explainable approaches can be useful. Current work has focused on \emph{black-box} explanation methods, which require no knowledge about the model internals, and analyze it as a black-box through input/response analysis. Methodologies such as LIME \cite{ribeiro2016should} provide local interpretation methods to explain single model decisions/predictions by linearizing a general model around the specific inputs, identifying the most relevant input features for that prediction. This approach guarantees considerable flexibility in the model selection. Other current black-box approaches include LEMNA \cite{Guo2018}, which extends LIME to improve local linearization, and SHAP \cite{SHAP2017}, which provides global interpretation methods based on aggregations. XAI is mainly linked to supervised learning scenarios and relies on domain knowledge to interpret the proposed explanations. 

Looking into XAI applications for unsupervised learning, and in particular, for clustering methodologies, very few solutions have been proposed. Authors of \cite{basak2005interpretable} propose the application of interpretable algorithms as a guideline for clustering execution. The authors propose a clustering methodology based on decision tree principles to create a rule path. Corral et al. \cite{corral2009explanations} develop a straightforward system to explain and describe the results of unsupervised learning for network security, only looking at the attributes common to most of the points in a cluster.

In this paper, we focus precisely on the problem of XAI to interpret unsupervised learning results. We introduce EXPLAIN-IT, a generic framework for unsupervised and self explainable learning, based on clustering. EXPLAIN-IT is not only capable to extract cohesive clusters from the data, but also to provide guidelines easing the interpretation of the results. In a nutshell, it uses clustering results as input to train a classification model, which is then explained through the application of black-box XAI approaches for interpretation of results. To serve as an example, we apply EXPLAIN-IT to the unsupervised analysis of QoE in YouTube video streaming, analyzing YouTube encrypted network traffic to autonomously identify different and relevant video resolution groupings caused by poor network performance. We note that EXPLAIN-IT is not a final system: this paper sets the initial steps into the overall ambitious goal so far described.

The remainder of the paper is organized as follows: Section \ref{xai_basics} provides additional concepts on the overall XAI domain, focusing on interpretability definitions and background. The EXPLAIN-IT framework is fully described in Section \ref{explainit_framework}. Section \ref{sec:evaluation} presents evaluation results on the capabilities of EXPLAIN-IT, using the YouTube video streaming QoE dataset described before. Finally, Section \ref{conclusion} provides concluding remarks, discussing on EXPLAIN-IT limitations and outlying future work.

\section{Explainable AI Basics}\label{xai_basics}

XAI refers to methods and techniques in the application of AI/ML such that the results of the solution can be understood by human experts \cite{gunning2017explainable, miller2019explanation, hazard2019, ribeiro2016should}. Achieving this goal is possible by identifying the inputs of a model leading to a particular output. XAI methods can be classified according to various criteria \cite{molnar2019}, namely: (i) intrinsic or \emph{post hoc} methodologies; (ii) objective of the interpretation method - (a) feature summary statistics, (b) feature summary visualization, (c) model internals or (d) data points; (iii) model-specific or model-agnostic; (iv) local or global interpretation.

Intrinsic interpretability means that the model is interpretable at its basis - e.g., linear models. Post hoc refers to the application of interpretation methods at testing time, after building the learning model. Different approaches can be used to interpret results, including: (ii.a) feature summary statistics, e.g., feature importance, (ii.b) feature summary visualization, e.g., partial dependence of features, (ii.c) model internals, e.g., checking the structure of the decision tree, (ii.d) data points, e.g., looking for representative prototypes of classes. The interpretability method can be model-specific if it applies only to one particular model, or model-agnostic - usually, post hoc interpretability is model-agnostic. At last, interpretability is defined global if it can explain the entire model behavior, or local if it explains an individual prediction.

Global, holistic model interpretability is frequently very hard to achieve, often because the model is too tangled, and because the input features usually correlate on a global scale.
This global picture can be more easily obtained at a modular level, combining local interpretability approaches. By zooming on a single instance and perturbing it, it is possible to obtain the local interpretability of the model. This approach allows the linearization of the problem and usually the neutralization of the dependencies between features, often allowing for more consistent results.

EXPLAIN-IT relies on local, post hoc, model-agnostic interpretability \cite{ribeiro2016should}, which provides high flexibility in terms of interpretable models, the obtainable explanations, and their representation. This overall flexibility allows a rapid switch between different problems, data types, and analysis models, making of EXPLAIN-IT a generic approach.

\begin{figure*}[t!]
\centering
\includegraphics[width=\textwidth]{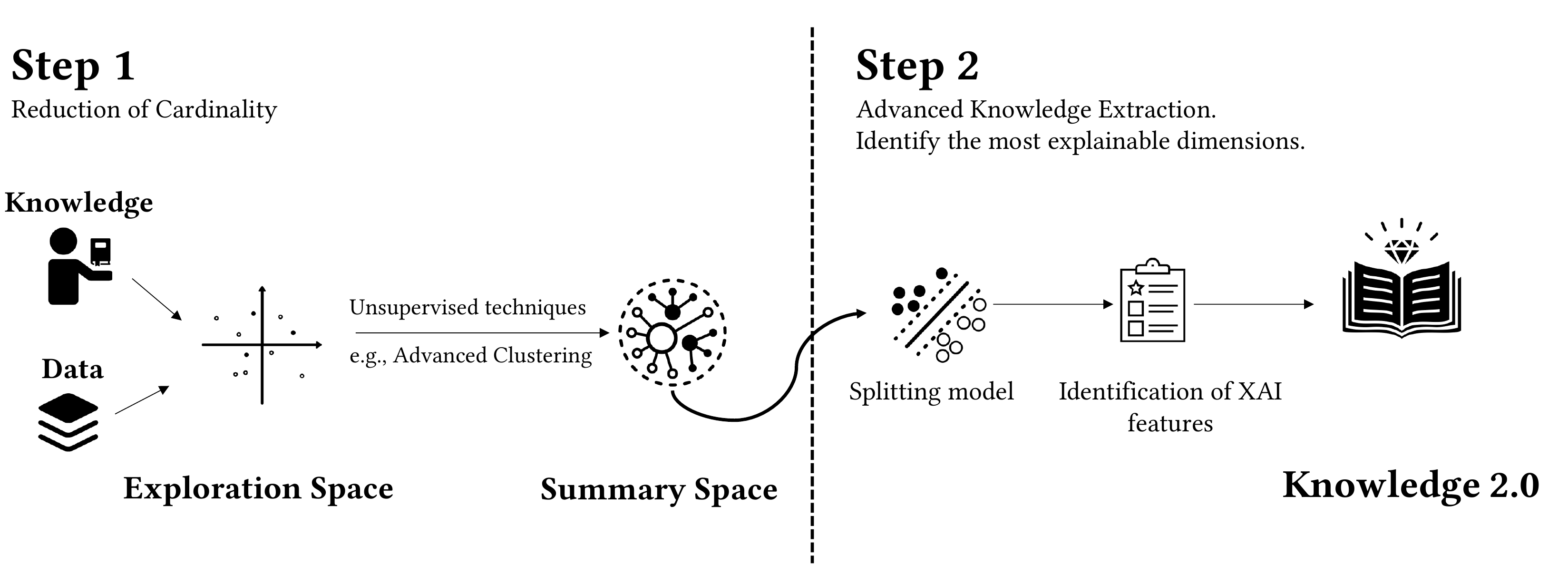}
\caption{The EXPLAIN-IT system. Data is firstly embedded into the \emph{exploration space}, relying on expert knowledge when available. The \emph{summary space} is the result obtained by clustering the exploration space. Next, we build a supervised data splitting model out of the clustering results. Finally, we apply an XAI approach (LIME) to this splitting model, interpreting the contents of the clusters by adding local interpretations.}
\label{fig:system_design}
\end{figure*}

\section{The EXPLAIN-IT Framework}\label{explainit_framework}

The goal of EXPLAIN-IT is to explain the outcome of unsupervised algorithms. In general terms, it addresses the process of knowledge discovery in datasets~\cite{fayyad_data_1996},  providing a comprehensive tool tackling the variety of steps of this process. Fig.~\ref{fig:system_design} depicts an overview of the system and its multiple steps. EXPLAIN-IT consists of two consecutive analysis steps, namely (i) reduction of the cardinality of the problem under analysis (e.g., data summarization and compression), and (ii) knowledge extraction and building of explanations.

\subsection{Exploration and Summary Spaces}

In the first step, EXPLAIN-IT relies (when available) on the knowledge of experts to extract the right features, structuring the data analysis and the way for interpreting the results. This process may include feature selection and engineering methodologies. The definition/extraction of features corresponds to embedding the data into an \emph{exploration space}, which serves as a basis for the data exploration process. Naturally, when no expert domain knowledge is available, the embedding would take place in an entirely blind fashion, using the features defined by the analyst. Once the data is embedded, EXPLAIN-IT uses unsupervised learning techniques to explore it, looking for relevant structures of interest. EXPLAIN-IT uses in particular clustering techniques, but any unsupervised methodology which creates a summary of the data can be applied. Clustering here plays the role of a meta-learning approach, which reduces the complexity of the analysis by aggregating similar instances, generating what we refer to as the \emph{summary space}. Different clustering algorithms can be used, depending on the desired result, amount of data, space dimensionality, and other constraints. For example, in case the user wants to obtain a certain number of groupings or classes, solutions that require the number of clusters as input are preferred, e.g., K-Means. Otherwise, if the data structure itself should determine the exploration process, or if anomaly detection is relevant in the process, other techniques such as density-based ones (e.g., DBSCAN) are better suited.

\subsection{Modeling the Summary Space}

The second step of EXPLAIN-IT consists of automatically characterizing the resulting clusters obtained in the summary space. A common way to measure the effectiveness of clustering algorithms \cite{iglesias2019}, it is to use intrinsic or derived characteristics, like complexity, steadiness, computational time, as well as internal and external validity metrics, like the silhouette and the rank indexes. Those measures are useful to evaluate the goodness of the algorithm, however, they provide little insight into what clusters contain. The main idea is, therefore, to identify, for each of the obtained clusters, the most relevant features explaining the assignment of each data instance to it. While this could be in-principle done by per-feature analysis and using weighted distances and linear discriminant functions, there is always a limitation of such an approach, based precisely on the considered notion of distance. Depending on which type of distance metric one would use, the obtained explanations would be different. For example, Euclidean distances would favor features defining spherical-like rules explaining the results, whereas sparse spaces and heavy-tailed distributions would impact probability-based or correlation-based distances. Also, global linear discrimination would not perform accurately in most practical cases. Other empirical approaches, such as manual inspection, relevant features description, or extraction of the most representative data instances, suffer from scalability, and the limitations mentioned above.

In this absence of a general solution to this problem, and inspired by the notions behind white-box and black-box XAI, we rely on a purely data-driven approach and decide to model the results of the clustering step through a supervised learning model, which we then explain through XAI. As we said before, to improve discrimination power, we do not rely on natively interpretable models (e.g., decision trees or linear discriminant functions), but we consider more powerful data splitting models. Here, in particular, we take Support Vector Machine (SVM) discriminant models \cite{vapnik1999overview}, which are we well-known for their performance to identify non-linear and complex boundaries among instances, relying on the use of kernel functions and the so-called ``kernel trick'' to construct such boundaries. 

\subsection{XAI with LIME}

The final step of the analysis consists of using black-box XAI approaches to finally interpret the structure of the summary space by explaining the SVM-based modeled clusters. In particular, we use LIME \cite{ribeiro2016should}, a model-agnostic interpretation approach which relies on local model linearization to identify the most relevant features leading to a particular decision, for an individual data instance. LIME relies on sampling for local model exploration and linearization. To explain the LIME approach, let us assume we have a complex model $f(\cdot)$, and a specific instance $x$ for which we want to explain the features leading to $f(x)$. LIME constructs a natively  interpretable model $g(\cdot)$, which is \emph{locally faithful} to $f(\cdot)$ in the vicinity of $x$, the latter captured by certain similarity measure $D_x(\cdot)$. For doing so, LIME randomly generates new instances $z$ around $x$, which are then weighted by $D_x(z)$ to define their local relevance. Finally, $g(\cdot)$ is built based on inputs $z$ and their corresponding labels $f(z)$. In particular, LIME uses linear discriminant functions to build $g(\cdot)$.

Naturally, other XAI methodologies such as SHAP~\cite{SHAP2017} are also very promising and could be part of EXPLAIN-IT. SHAP has solid mathematical foundations, but it has a much higher complexity that makes it hard to use on exploratory approaches. However, it results in better explanations - or at least approximations, making it more reliable in case of more stringent requirements. Nevertheless, LIME offers the best trade-off between computational time and interpretability, therefore its high popularity.

By combining the explanations provided by LIME for each data instance belonging to each cluster, EXPLAIN-IT finally provides guidelines easing the interpretation of the clustering results. The overall solution is very modular, allowing different settings and flexibility in the various stages of the process.

\section{EXPLAIN-IT for YouTube QoE}\label{sec:evaluation}

To serve as an application example of EXPLAIN-IT, we apply the proposed system to the unsupervised analysis of QoE in YouTube video streaming, analyzing YouTube encrypted network traffic to autonomously identify different and relevant video resolution groupings caused by poor network performance. The main challenge in the analysis of YouTube QoE from in-network measurements is that the wide adoption of end-to-end encryption through HTTPS blinds previous Deep Packet Inspection (DPI) based approaches, making of ML an appealing solution. We have extensively worked on this problem \cite{seufert2019infocom, seufert2019icin}, but always considering a supervised learning approach. For this example, we would focus exclusively on the prediction and analysis of the average video resolution or video quality (AVGQ) of YouTube video sessions as the target metric.

Next, we describe the YouTube dataset analyzed in this AVGQ prediction task, and go across each of the steps followed by EXPLAIN-IT, including the clustering performance and its validation through standard metrics, the modeling of the obtained clusters through SVM, and the final obtained explanations through LIME.

\begin{table*}[t!]
\caption{Supervised and unsupervised quality metrics for clustering. The higher the values, the better the results.}
\resizebox{\textwidth}{!}{
\begin{tabular}{lrrrrrrr}
        \toprule
        algorithm & adj\_mutual\_info &  adj\_rand &                         completeness &  fowlkes\_mallows &  homogeneity &  silhouette &  v\_measure \\
        \midrule
        Agglomerative\_Ward & 0.131581 &             0.103339 &            0.217531 &               0.532475 &           0.131746 &        \textbf{0.379734} &         0.164104 \\
        Agglomerative\_Single & 0.034626 &             0.026033 &            0.197189 &               \textbf{0.581536} &           0.034815 &          0.077997 &         0.059182 \\
        KMeans & \textbf{0.191741} &             \textbf{0.190060} &            \textbf{0.283335} &               0.557967 &          \textbf{ 0.191894} &          0.329705 &         \textbf{0.228817} \\
        Birch & 0.132633 &             0.109831 &            0.224226 &               0.535183 &           0.132798 &          0.325101 &         0.166805 \\
        \bottomrule
\end{tabular}}
\label{tab:cl_metrics_all_features}
\end{table*}

\begin{figure}[t!]
    \centering
    \includegraphics[width=\columnwidth]{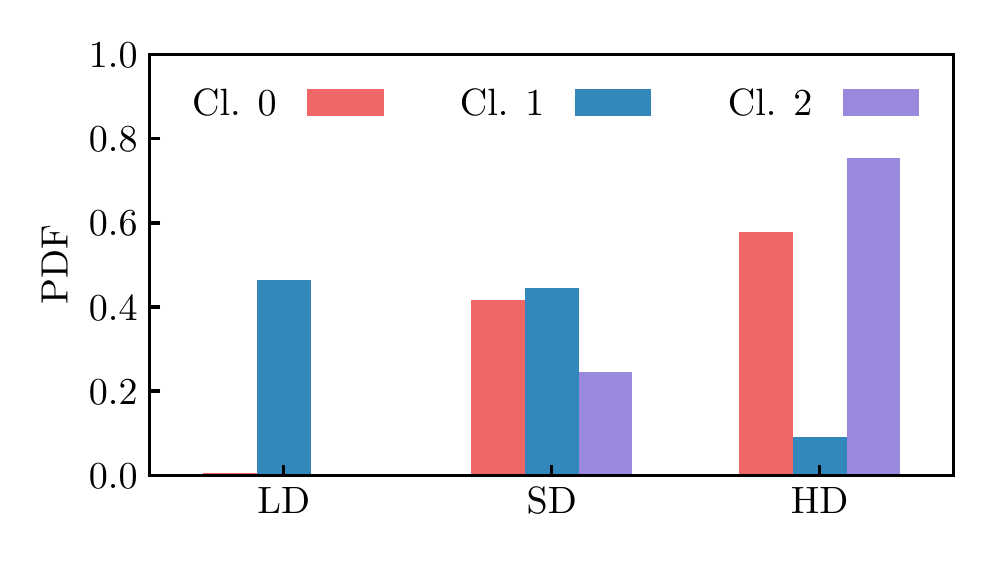}
    \caption{Distribution of real labels - LD, SD, and HD in the clusters generated by Agglomerative\_Ward.}
    \label{fig:labels_distribution}
\end{figure}

\begin{figure*}[t!]
    \centering
    \begin{subfigure}{0.33\textwidth}
        \centering
        \includegraphics[width=.95\linewidth]{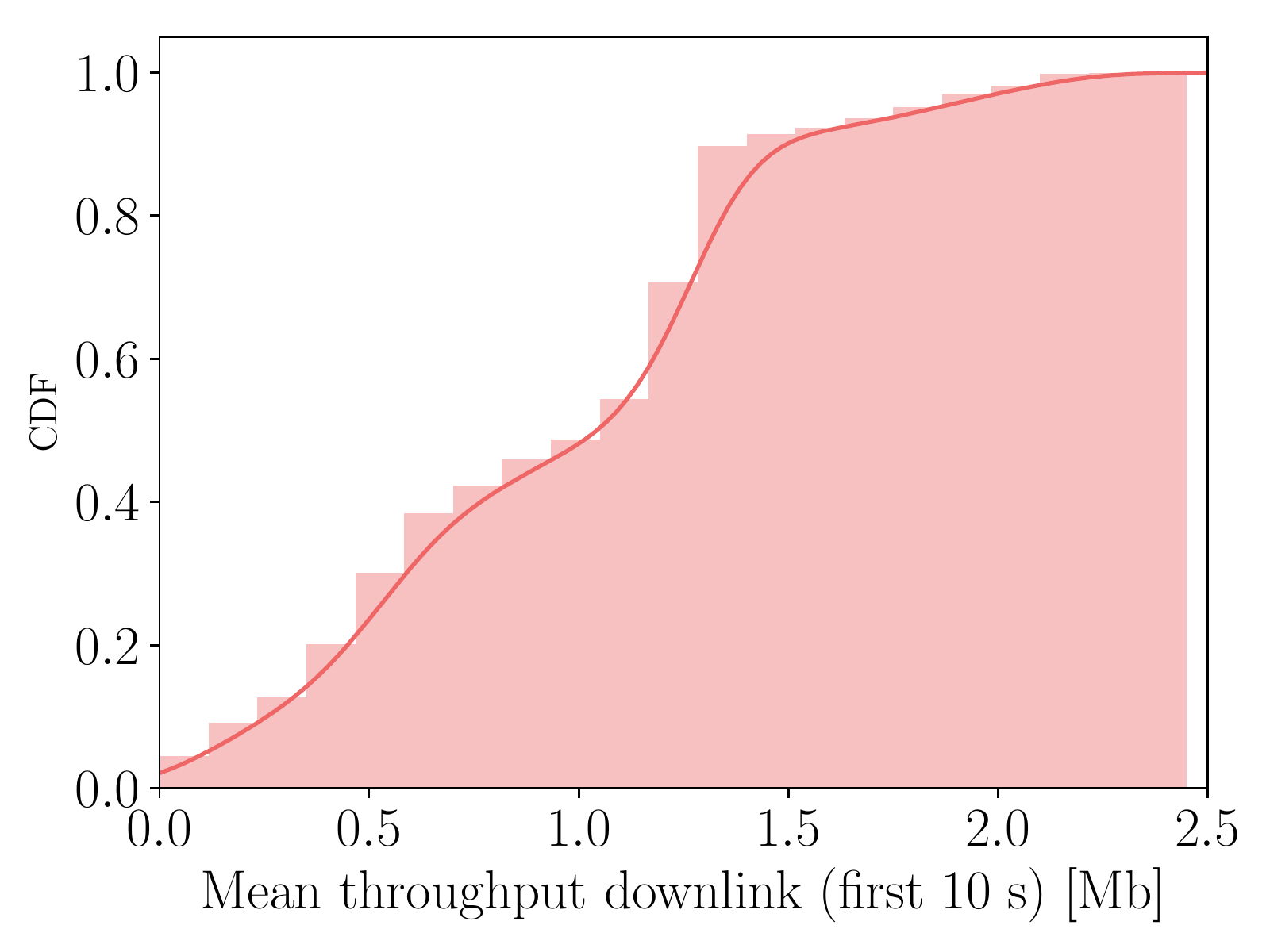}
        \caption{YouTube sessions in Cluster 0}
         \label{fig:throughput_cl0}
    \end{subfigure}
    \begin{subfigure}{0.33\textwidth}
        \centering
        \includegraphics[width=.95\linewidth]{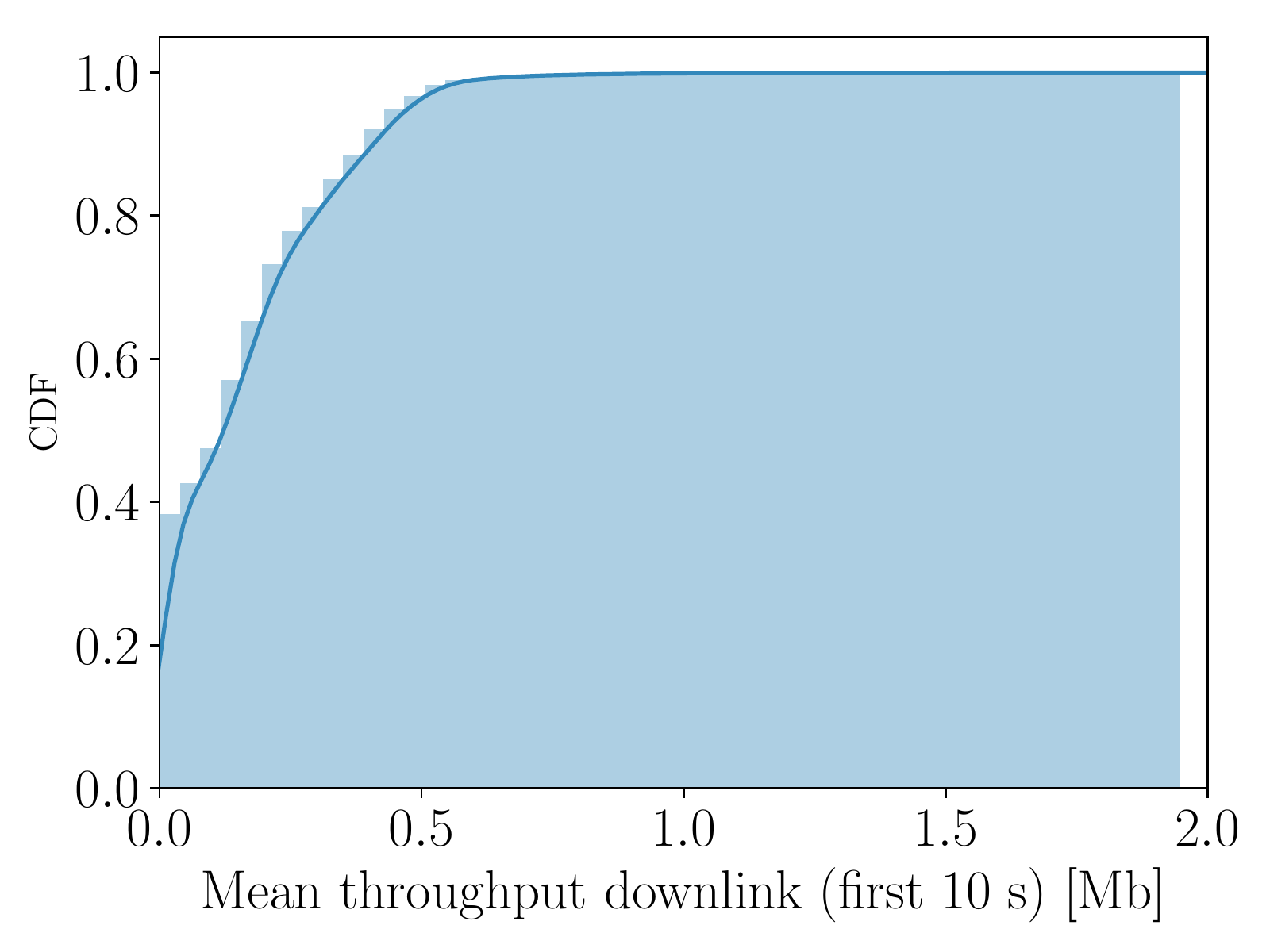}
        \caption{YouTube sessions in Cluster 1}
        \label{fig:throughput_cl1}
    \end{subfigure}
    \begin{subfigure}{0.33\textwidth}    
        \centering
        \includegraphics[width=.95\linewidth]{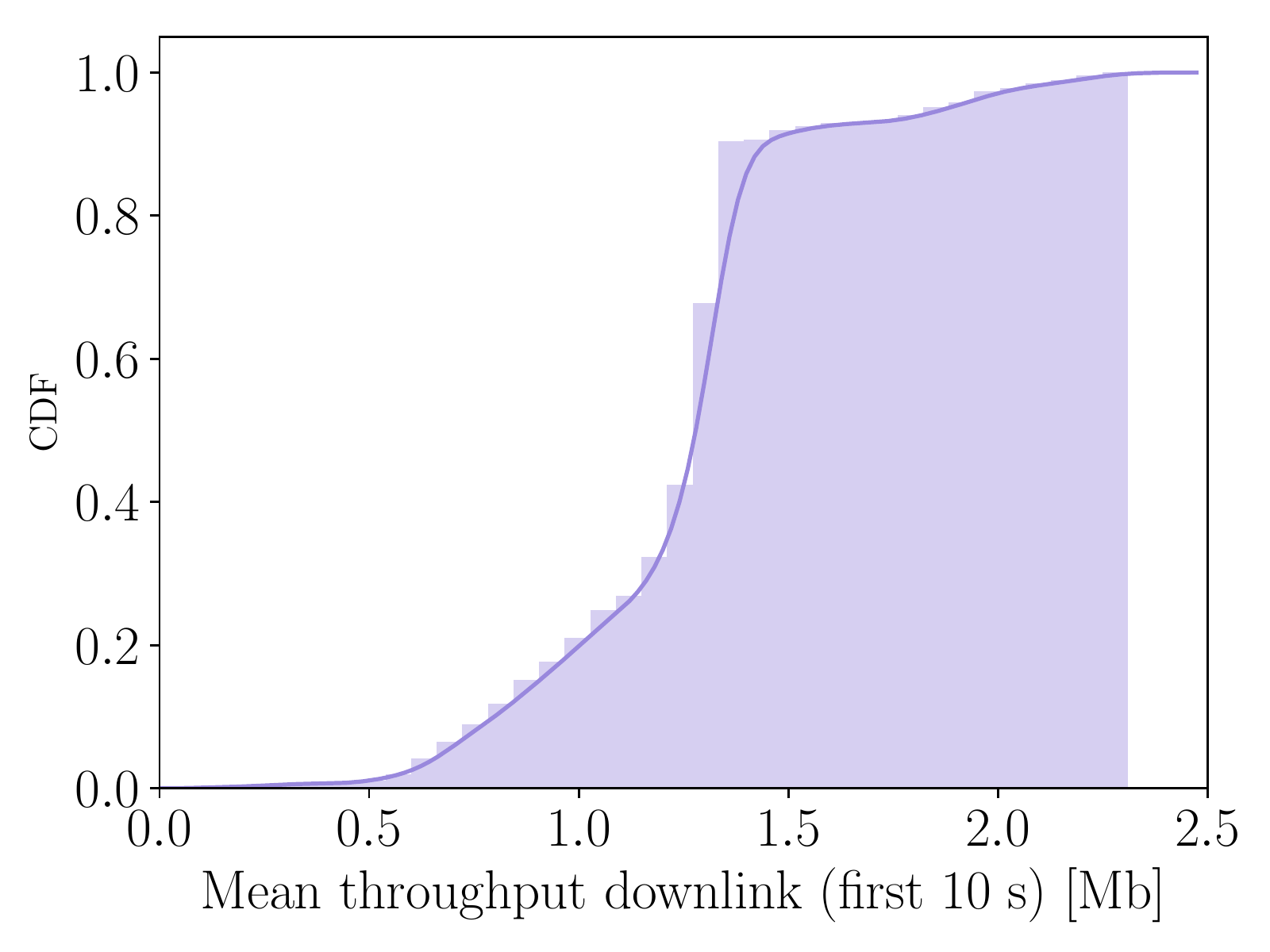}
        \caption{YouTube sessions in Cluster 2}
        \label{fig:throughput_cl2}
    \end{subfigure}
    \caption{Average downlink throughput (ADT) distribution on the first 10s slot, per cluster. While there is strong variance within each cluster, ADT distributions are ordered, with ADT($C_1$) < ADT($C_0$) < ADT($C_2$).}
    \label{fig:throughput_benchmarking}
\end{figure*}

\subsection{YouTube QoE Dataset}

The dataset we analyze consists of 10.654 YouTube video sessions, collected in \cite{seufert2019infocom, seufert2019icin} from different sources, including smartphones (HTML player and YouTube app) and desktop (HTML player) devices, and considering both TCP and QUIC protocols. A set of $m = 477$ features, extracted from the encrypted stream of packets, describes each video session. These include features at the full video session level (e.g., session downlink throughput) as well as at different time resolutions, with time slots aggregating packets in $\Delta t = [1, 5, 10]$ seconds. Features include not only traditional metrics such as min/avg/max values, but also sampled values of the observed empirical distributions, including percentiles as well as dispersion metrics such as entropy $h(\cdot)$. Several QoE-related metrics are directly measured at the player side and considered as ground-truth metrics, including the number and duration of stalling events, number of quality switches, average video bitrate, and AVGQ. See \cite{seufert2019infocom} for a full description of features and labels.

Using AVGQ as target metric, we build a discrete label to analyze the problem as a three classes classification task, including: Low Definition (LD) - $AVGQ < 480$, Standard Definition (SD) - $480 \leq AVGQ \leq 720$, and High Definition (HD) - $AVGQ \geq 720$. We use these labels as ground truth for validation purposes only, and not as part of the EXPLAIN-IT processing itself. We use the full set of 477 features as the embedding, exploration space.

\subsection{Clustering Approaches and Validation}

We evaluate four different algorithms in the clustering step. Our goal is to find three clusters representing the apriori expected AVGQ classes - LD, SD, and HD. Therefore, our selection of algorithms includes only those methodologies that have the number of clusters as an input parameter. The selected algorithms include: \textit{Agglomerative\_Ward} - hierarchical clustering with Ward Links (Ward minimizes the variance of the clusters during the merging phase); \textit{Agglomerative\_Single} -  hierarchical clustering with Single Links (Single uses the minimum of the distances between all observations of the two sets); \textit{K-Means}: best known clustering algorithm, with K=3; \textit{BIRCH} - Balanced Iterative Reducing and Clustering using Hierarchies \cite{zhang1996birch}. We use the implementations of these algorithms, as provided by Scikit-Learn. 

We begin by evaluating the performance of these clustering algorithms using popular and well-accepted validation metrics, namely adjusted mutual information, adjusted random score, completeness, Fowlkes-Mallows, homogeneity, silhouette, and V-measure, see \cite{iglesias2019} for specific definitions on these metrics. For all these metrics, the higher the value, the better the results. Tab.~\ref{tab:cl_metrics_all_features} depicts the obtained values for these metrics, for the different clustering algorithms. K-Means and Agglomerative\_Ward have, on average, the best results over the considered metrics. Based on the results for the silhouette score, we select \textit{Agglomerative\_Ward} as the clustering algorithm for the next step. The silhouette score is a measure of how similar an instance is to the other cluster neighboring instances as compared to instances in other clusters. 

\begin{figure}[t!]
    \centering
    \includegraphics[width=\columnwidth]{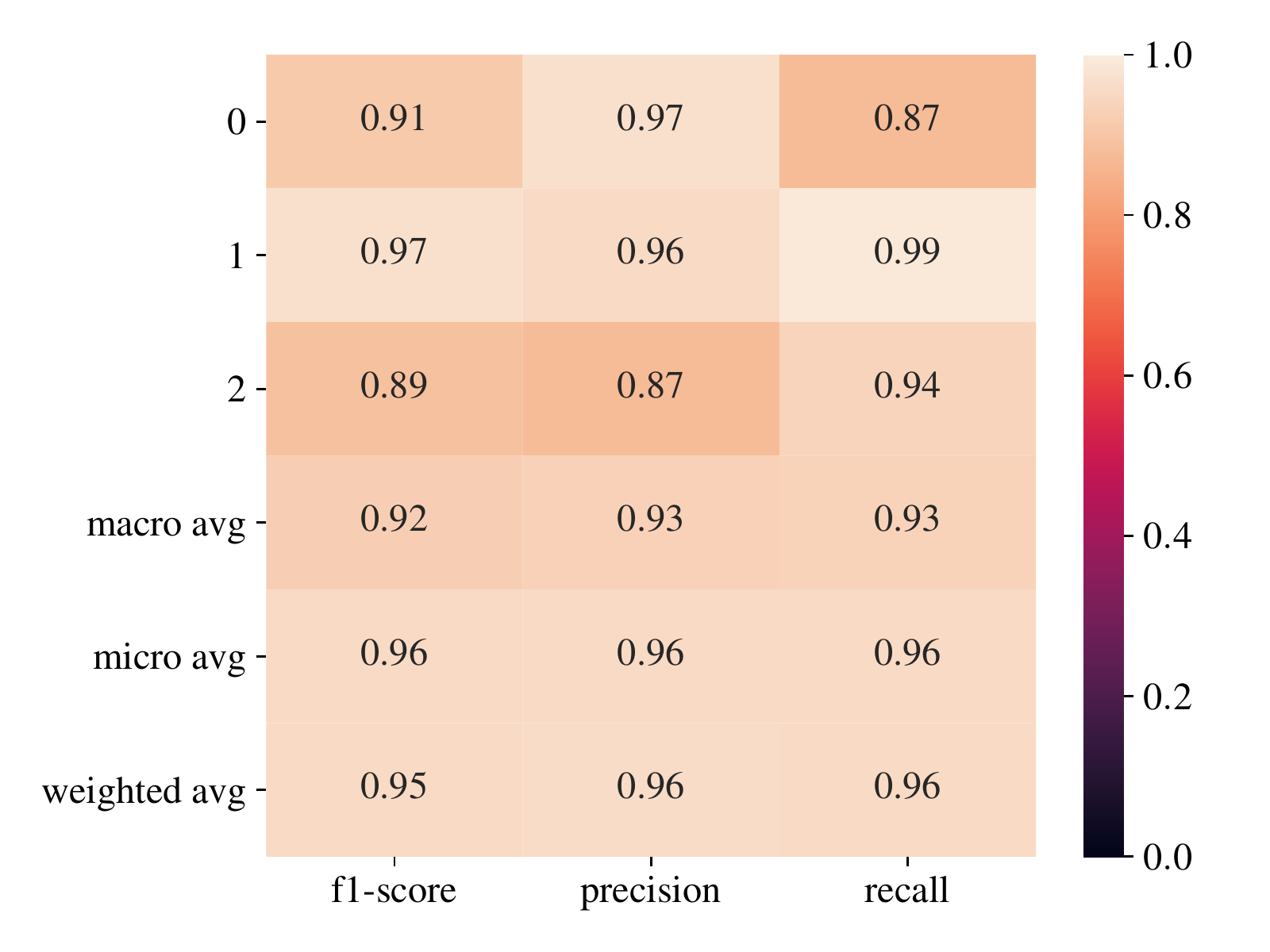}
    \caption{10-fold cross SVM classification results, using Agglomerative\_Ward clusters as labels.}
    \label{fig:aggl_svm-classification-report}
\end{figure}

We are taking the results provided by \textit{Agglomerative\_Ward}, Fig.~\ref{fig:labels_distribution} to show how the distribution of the clustered sessions among the three different AVGQ labels. Even if the split of the video sessions is not pure, we can identify three different groupings. Cluster $C_1$ contains mostly LD and SD YouTube sessions; cluster $C_0$ is more shifted towards SD and HD, while HD sessions predominantly characterize cluster $C_2$. To further verify these observations, we study the distribution of self-explainable, domain expert knowledge features across clusters. In particular, it is coherent to accept that higher resolution video sessions would correlate to sessions with higher Average Downlink Throughput (ADT), and especially when considering the beginning of the video session. Fig.~\ref{fig:throughput_benchmarking} depicts the distribution of ADT values over the first 10 seconds slot of the video sessions. Again, we can identify the trend for the three classes: $C_1$ shows the lowest values of ADT, with more than 95\% of the sessions having an ADT < 0.5 Mbps. For video sessions in $C_0$, ADT is higher than 1 Mbps for about 60\% of the sessions, whereas this amount increases to about 80\% of the sessions in $C_2$.

This type of explanation is what one could expect by manually inspecting each of the most relevant and discriminative features. However, manually inspecting the full set of 477 features is not doable in practice, therefore the relevance of automating this process through EXPLAIN-IT.

\subsection{Clusters' Modeling with SVM}

We want to use a model that interprets the clustering results. We want to understand why the algorithm assigned an instance to a specific cluster and which are the main features that are distinctive for that assignment. As explained before, we do this by training an SVM model on the clustering results of Agglomerative\_Ward, i.e., using the resulting clusters as labels for the classification algorithm. Fig.~\ref{fig:aggl_svm-classification-report} depicts the performance achieved by SVM in this classification task in terms of precision, recall, and f-score - per class and micro/macro/weighted average, relying on 10-fold cross-validation. As observed, the SVM model is capable of almost entirely model the clusters obtained by Agglomerative\_Ward, with f-scores close or above 90\% for all three clusters. The advantage now is that we have a model which represents the obtained cluster partitions as classification classes, and can be automatically used to interpret the assignments of instances to each of these clusters.

\subsection{Explaining Results with LIME}

The last step of EXPLAIN-IT is to use LIME to interpret the cluster assignment decisions as modeled by the SVM classifier. We apply LIME with a neighborhood of 100 random instances for each YouTube session to be explained. As output, LIME provides the probability of this session belonging to a specific class - in this case, modeling a cluster, and a list of the most relevant features and associated values leading to the decision. We report an example in Tab.~\ref{tab:explainable_AI}, which shows the top 10 features that characterize the prediction of a YouTube session as belonging to cluster $C_2$. As expected, and based on the results observed in Fig. \ref{fig:labels_distribution} and Fig. \ref{fig:throughput_benchmarking}, we see that the explanations provided by LIME point to multiple features, which should be higher than certain thresholds to belong to $C_2$. For example, packet length in the downlink direction should be higher than 1,380 bytes, uplink accumulated number of bytes sent at the initial time slots of 1 second should be higher than 10 to 20 KB, downlink accumulated number of bytes sent at time slots of 5 seconds should be higher than 2.7 MB, and so on. These explanations show that i) some features drive the decision better than others; ii) these features relate to the quality of the connection and content of the video session; iii) top percentiles of the distributions are very relevant for the assignments, suggesting that peak values are more important than averages.

By aggregating the explanations provided for single instances, we can further gain insights about the generated clusters and their commonalities/differences. For example, Tab.~\ref{tab:explainable_AI_common} shows the intersection among the 30 most popular features deciding the assignment to clusters $C_1$ and $C_2$, considering all the video sessions in each cluster. As we see, clusters $C_1$ (mostly LD and SD, cf. Fig \ref{fig:labels_distribution}) and $C_2$ (mostly SD and HD, cf. Fig \ref{fig:labels_distribution}) share six features which influence the assignment of video sessions to them, but with different relevance, and also having different value intervals. The intersection comes mostly from the SD video sessions, spread among the three clusters.

\begin{table}[t!]
    \caption{EXPLAIN-IT reported features for an instance assigned to $C_2$. The table reports the features and their importance in the decision process.}
    \centering
    \resizebox{\columnwidth}{!}{
    \begin{tabular}{lr}
    \toprule
    \textbf{Feature Interval} & \textbf{Feature Importance} \\
    \midrule
    uplink\_bytes\_second\_slot\_1s > 10,470 & 0.100 \\
    dist\_packet\_length\_downlink\_p25 > 1,380 & 0.090 \\
    dist\_slotted\_uplink\_bytes\_p97\_1s > 18,446 & 0.083 \\
    uplink\_packets\_first\_slot\_5s > 860 & 0.070 \\
    420,630 < dist\_slotted\_bytes\_p97\_1s <= 902,384 & 0.067 \\
    dist\_slotted\_downlink\_bytes\_p97\_5s > 2,711,880 & 0.059 \\
    dist\_slotted\_downlink\_bytes\_h\_1s > 0.66 & 0.053 \\
    335.405 < dist\_slotted\_uplink\_packets\_p99\_1s <= 502 & 0.044 \\
    dist\_slotted\_uplink\_bytes\_p90\_1s > 7,630 & 0.044 \\
    dist\_slotted\_bytes\_mean\_5s > 845,018 &  0.037 \\
    \bottomrule
    \end{tabular}}
    \label{tab:explainable_AI}
\end{table}

\section{Concluding Remarks}\label{conclusion}

EXPLAIN-IT is a novel approach to \textbf{support} unsupervised data analysis with improved insights on the obtained results. It may complement and, in some cases, even substitute traditional clustering evaluation methodologies. The overall target for EXPLAIN-IT is to implement a fully end-user-interactive unsupervised analysis system, which the end user could apply to refine the analysis and obtain better insights step by step. For example, the end-user could use the results of an EXPLAIN-IT round to identify and select the most meaningful and explainable features to re-use in the next step, refining the embedding on the exploration space.   

The application of EXPLAIN-IT to the unsupervised analysis of YouTube QoE shows that it is possible to improve the interpretation of clustering results by relying on XAI principles. However, being unsupervised by nature, EXPLAIN-IT has still many open questions that we are trying to solve. For example, the cluster modeling step through supervised learning inherently introduces a bias due to the application of a specific model, in our case, SVM. In terms of XAI, LIME works by generating random instances to extract the explanations. However, these random instances can turn out to be meaningless. Especially if the data itself is complex, and with many features, the explanations obtained by LIME in such a way can likely change. We are working on these and other open issues to improve EXPLAIN-IT. For example, we are currently testing more robust XAI approaches such as SHAP and LEMNA to improve explainability properties, as well as working on different unsupervised models - such as auto-encoders, as well as other supervised ones for clustering-results-modeling - such as CNNs. We are also evaluating to use the clustering results themselves as the model to explain, avoiding the bias introduced by the supervised-based modeling step. As we said before, this has associated limitations and challenges we should tackle to provide a significant advantage in terms of generalization of the framework.   

\begin{table}[t!]
    \caption{Common ``most popular'' features explaining assignments to $C_1$ and $C_2$.}
    \centering
    \resizebox{\columnwidth}{!}{
    \begin{tabular}{c|c|c}
    \toprule
    \textbf{Ranking in $C_1$} & \textbf{Feature Name} & \textbf{Ranking in $C_2$}\\
    \midrule
    2 & uplink\_throughput\_second\_slot\_1s & 16\\
    4 & uplink\_packets\_second\_slot\_1s & 23\\
    5 & uplink\_bytes\_second\_slot\_1s & 10\\
    6 & downlink\_packets\_second\_slot\_1s & 5\\
    7 & downlink\_throughput\_second\_slot\_1s & 11\\
    9 & downlink\_bytes\_second\_slot\_1s & 9\\
    \bottomrule
    \end{tabular}}
    \label{tab:explainable_AI_common}
\end{table}

\balance

\bibliographystyle{ACM-Reference-Format}
\bibliography{references}

\end{document}